\documentclass{article}

\usepackage[preprint]{corl_2026} % Uncomment for pre-prints (e.g., arxiv).

\usepackage{microtype}
\usepackage{graphicx}
\usepackage{subcaption}
\usepackage{booktabs} % for professional tables

\usepackage{textcomp} % for \textdegree
\usepackage{colortbl} % for \cellcolor

\usepackage{amsmath}
\usepackage{amssymb}
\usepackage{mathtools}
\usepackage{amsthm}
\usepackage{enumitem} % for compact lists
\usepackage{wrapfig} % for text-wrapping around tables/figures

\usepackage[capitalize,noabbrev]{cleveref}

\theoremstyle{plain}

\theoremstyle{definition}

\theoremstyle{remark}

\usepackage[disable,textsize=tiny]{todonotes}

\title{NavOne: One-Step Global Planning for Vision-Language Navigation on Top-Down Maps}

\author{
  Dijia Zhan$^*$, Jinyi Li$^*$, Chenxi Zheng, Shaoyu Huang, Yong Li$^\dagger$, Jie Tang$^\dagger$, Xuemiao Xu$^\dagger$\\
  South China University of Technology\\
  {\small $^*$Equal contribution \quad $^\dagger$Corresponding authors}\\
}

\begin{document}

\maketitle

\begin{abstract}
  Existing Vision-Language Navigation (VLN) methods typically adopt an egocentric, step-by-step paradigm, which struggles with error accumulation and limits efficiency. While recent approaches attempt to leverage pre-built environment maps, they often rely on incrementally updating memory graphs or scoring discrete path proposals, which restricts continuous spatial reasoning and creates discrete bottlenecks. We propose Top-Down VLN (TD-VLN), reformulating navigation as a one-step global path planning problem on pre-built top-down maps, supported by our newly constructed R2R-TopDown dataset. To solve this, we introduce NavOne, a unified framework that directly predicts dense path probabilities over multi-modal maps in a single end-to-end forward pass. NavOne features a Map Fuser for early modality integration, and extends Attention Residuals for spatial-aware depth mixing. Extensive experiments on R2R-TopDown show that NavOne achieves state-of-the-art performance among map-based VLN methods, with a planning-stage speedup of $8\times$ over existing map-based baselines and $80\times$ over egocentric methods, enabling highly efficient global navigation. See \href{https://altman-conquer.github.io/NavOnePage/}{\texttt{https://altman-conquer.github.io/NavOnePage/}} for additional project details.
\end{abstract}


\vspace{-1.0em}

\begin{figure}[h]
\centering
\includegraphics[width=\linewidth]{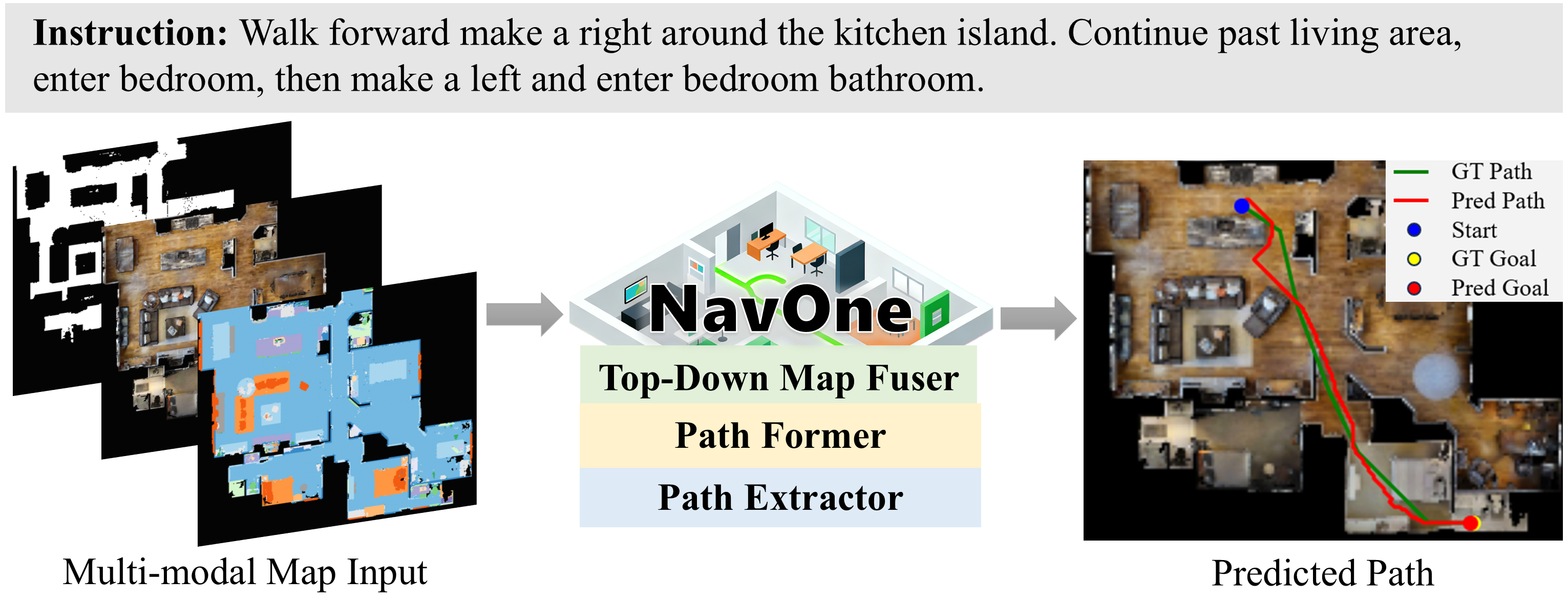}
\caption{Overview of NavOne. Given a language instruction and multi-modal top-down map inputs (RGB, occupancy, and semantic layers), NavOne processes them through three components—\textbf{Top-Down Map Fuser}, \textbf{Path Former}, and \textbf{Path Extractor}—to predict a complete navigation path in a single forward pass. The predicted path (red) and goal (red dot) closely follow the ground-truth trajectory (green) and goal (yellow dot).}
\label{fig:overview}
\end{figure}

\vspace{-1.5em}

\section{Introduction}

Vision-Language Navigation (VLN) \cite{anderson2018vision,qi2020reverie} is a core problem in embodied AI, requiring agents to interpret natural language instructions and navigate physical environments. Existing methods largely adopt an egocentric, step-by-step decision-making paradigm \cite{banerjee2021robotslang,krantz2020beyond}, iteratively selecting actions based on local visual observations and partial state histories \cite{chen2021history,chen2022think}. This formulation has been extended through graph-based planning, memory mechanisms, pre-training strategies, and continuous 3D environments \cite{zhang2024vision}, achieving strong performance on established benchmarks.

Despite steady progress, egocentric and sequential formulations continue to face fundamental challenges in long-horizon navigation \cite{song2025towards}: errors compound over time, global spatial structure is weakly modeled, and inference requires repeated action prediction at each timestep, resulting in substantial computational cost in large environments \cite{zhang2024vision}.

These limitations, however, are not fundamental. A key observation is that many real-world robots already operate with access to full or partial environment maps, obtained via pre-exploration using standard SLAM pipelines \cite{labbe2019rtab}. Recent map-based VLN methods attempt to exploit this by explicitly building memory graphs to track sequential exploration \cite{zhang2025mapnav} or by generating and scoring discrete path proposals \cite{wang2025instruction}. However, reliance on incrementally updated, discrete pipelines limits continuous spatial reasoning and remains computationally bottlenecked by candidate generation. A more principled approach would directly predict a complete navigation path over the full map in a single forward pass, jointly optimizing all components end-to-end.

We introduce \textbf{T}op-\textbf{D}own \textbf{VLN} (\textbf{TD-VLN}), reformulating navigation as a \emph{one-step global planning problem} on top-down maps, where an agent must predict a complete navigation path from a single top-down map observation given a language instruction, bypassing the need for step-by-step actions or discrete proposals. To support this, we construct the \textbf{R2R-TopDown} dataset, providing multi-modal map representations (RGB, occupancy, semantics) paired with instructions and trajectories.

To this end, we propose \textbf{NavOne} (\Cref{fig:overview}), a unified end-to-end framework that generates complete navigation paths in a single forward pass. NavOne consists of: 1) a \textbf{Top-Down Map Fuser} for joint multi-modal map representation; 2) a \textbf{Path Former} encoder-decoder that integrates language, pose, and spatial inputs to predict dense path and goal distributions; and 3) a \textbf{Path Extractor} that converts predictions into executable trajectories. By eliminating iterative action prediction, NavOne achieves a planning-stage speedup of $8\times$ over existing map-based baselines and $80\times$ over egocentric methods, enabling highly efficient global navigation. Ablation studies further validate the benefits of spatial-aware depth queries for position-dependent feature mixing.

\section{Related Work}
\label{sec:related}

\textbf{Vision-Language Navigation.}
Vision-Language Navigation (VLN) traditionally maps language to sequential egocentric actions \cite{anderson2018vision, an2024etpnav, wang2019reinforced, krantz2020beyond, krantz2021waypoint}. Despite advances \cite{Gao2023FastSlowTA}, the dominant continuous decision-making formulation remains computationally expensive and prone to compounding errors over long horizons \cite{zhang2024vision,szot2024grounding, chen2025affordances}: errors accumulate at each decision step \cite{wang2025think}, global spatial structure is weakly modeled \cite{an2023bevbert}, and repeated action prediction incurs substantial cost in large environments \cite{kang2025harnessing}. We circumvent these issues by reformulating VLN as a one-step global planning problem when top-down maps are available.

\textbf{Map-Based and Top-Down Navigation.}
Top-down maps provide spatial priors to improve generalization, interpretability, and long-horizon reasoning \cite{zhaolearning, li2024semantic, hong2023learning, feng2025vpn, zhong2024topv}. Recent advances in indoor mapping and robotic perception have made rich multi-modal spatial representations increasingly accessible, including RGB maps, occupancy grids, and semantic maps \cite{georgakis2022cross, chaplot2020neural}. Related efforts in cross-modal map learning \cite{georgakis2022cross} and neural topological representations \cite{chaplot2020neural} demonstrate the value of structured spatial knowledge for navigation. While previous works often use maps as auxiliary cues or hierarchical guides with discrete proposal bottlenecks \cite{wang2025instruction}, or build incrementally updated memory graphs \cite{zhang2025mapnav}, TD-VLN places them at the center to directly predict complete, dense navigation paths in a single forward pass.

% \textbf{World Models for Navigation.}
% World models learn latent representations of environment dynamics \cite{Ha2018RecurrentWM,lecun2022path}, enabling agents to simulate future states for planning and decision-making \cite{hafner2019learning}. Recurrent State-Space Models (RSSM) and the Dreamer family \cite{hafner2019dream} demonstrated the effectiveness of imagination-based planning in reinforcement learning \cite{hafner2023mastering}. In navigation, PathDreamer \cite{koh2021pathdreamer} introduced visual world models to synthesize future observations at candidate viewpoints, while DreamWalker \cite{wang2023dreamwalker} applied heuristic search over imagined egocentric views in continuous VLN. Navigation World Models \cite{bar2025navigation} further advanced controllable video generation for trajectory imagination, and NavMorph \cite{yao2025navmorph} proposed self-evolving world models with test-time adaptation. However, existing navigation world models primarily operate in egocentric space and rely on sequential action prediction, which incurs significant computational cost over long horizons. In contrast, our approach explores a complementary direction by performing one-step global planning on top-down maps, enabling efficient and interpretable path inference.

\textbf{Environment Map Construction via Pre-Exploration.}
Rich top-down environment maps can be acquired in a one-time pre-exploration phase using modern SLAM pipelines, with frontier-based exploration~\cite{yamauchi1997frontier} providing a principled strategy for systematic coverage. Traditional 2D LiDAR SLAM systems such as Cartographer~\cite{hess2016real} and RTAB-Map~\cite{labbe2019rtab} produce stable occupancy grids via pose graph optimization and loop closure; LiDAR-inertial approaches including LIO-SAM~\cite{shan2020liosam} and FAST-LIO2~\cite{xu2022fastlio2} further improve trajectory accuracy in medium-scale indoor environments. More recently, neural implicit SLAM~\cite{zhu2022niceslam} and 3D Gaussian Splatting SLAM~\cite{keetha2024splatam} enable photorealistic dense reconstruction, though their 3D outputs typically require additional projection to yield 2D navigable maps.
% Top-down maps can be acquired once via frontier-based exploration~\cite{yamauchi1997frontier} with modern SLAM pipelines~\cite{hess2016real,labbe2019rtab,shan2020liosam}, which produce stable occupancy grids; recent neural implicit and Gaussian Splatting SLAM methods~\cite{zhu2022niceslam,keetha2024splatam} yield denser reconstructions at the cost of additional 3D-to-2D projection.

% \textbf{Attention Residuals.}
% Standard Transformers~\cite{vaswani2017attention} employ additive residual connections that restrict each layer to refining only its immediate predecessor. Attention Residuals~\cite{team2026attention} replaces this fixed skip connection with a depth-wise attention mechanism that allows each layer to selectively mix all previous layer outputs via learned queries and softmax weighting. We extend this to navigation by introducing spatial-aware depth queries that enable position-dependent layer mixing, allowing different map regions (e.g., obstacle boundaries vs.\ open spaces) to attend to the most informative feature hierarchy.

\section{R2R-TopDown Dataset}
\label{sec:dataset}

% We transfer the VLN dataset R2R-CE to the TD-VLN task by replacing egocentric observations with top-down map representations, building a new dataset R2R-TopDown for TD-VLN. Each scene from Matterport3D (MP3D) environments is represented by three types of 2D top-down maps: an RGB map, an occupancy map indicating navigable areas, and a semantic map with 41 object categories. Detailed construction procedures are provided in Appendix~\ref{appendix:dataset-construction}. \Cref{fig:dataset-examples} shows representative examples of these multi-modal map inputs.

We construct the R2R-TopDown dataset for TD-VLN by transferring R2R-CE trajectories to top-down maps. Each scene from Matterport3D (MP3D) environments is represented by three types of 2D top-down maps: an RGB map, an occupancy map indicating navigable areas, and a semantic map with 41 object categories. Detailed construction procedures are provided in Appendix~\ref{appendix:dataset-construction}. \Cref{fig:dataset-examples} shows representative examples of these multi-modal map inputs.

% We project 3D navigation trajectories from R2R-CE onto 2D map planes, filter to retain only single-floor episodes, and include agent initial state (position and orientation) for each episode.

\begin{figure}[h]
\centering
\setbox5=\hbox{\includegraphics[width=0.5\linewidth]{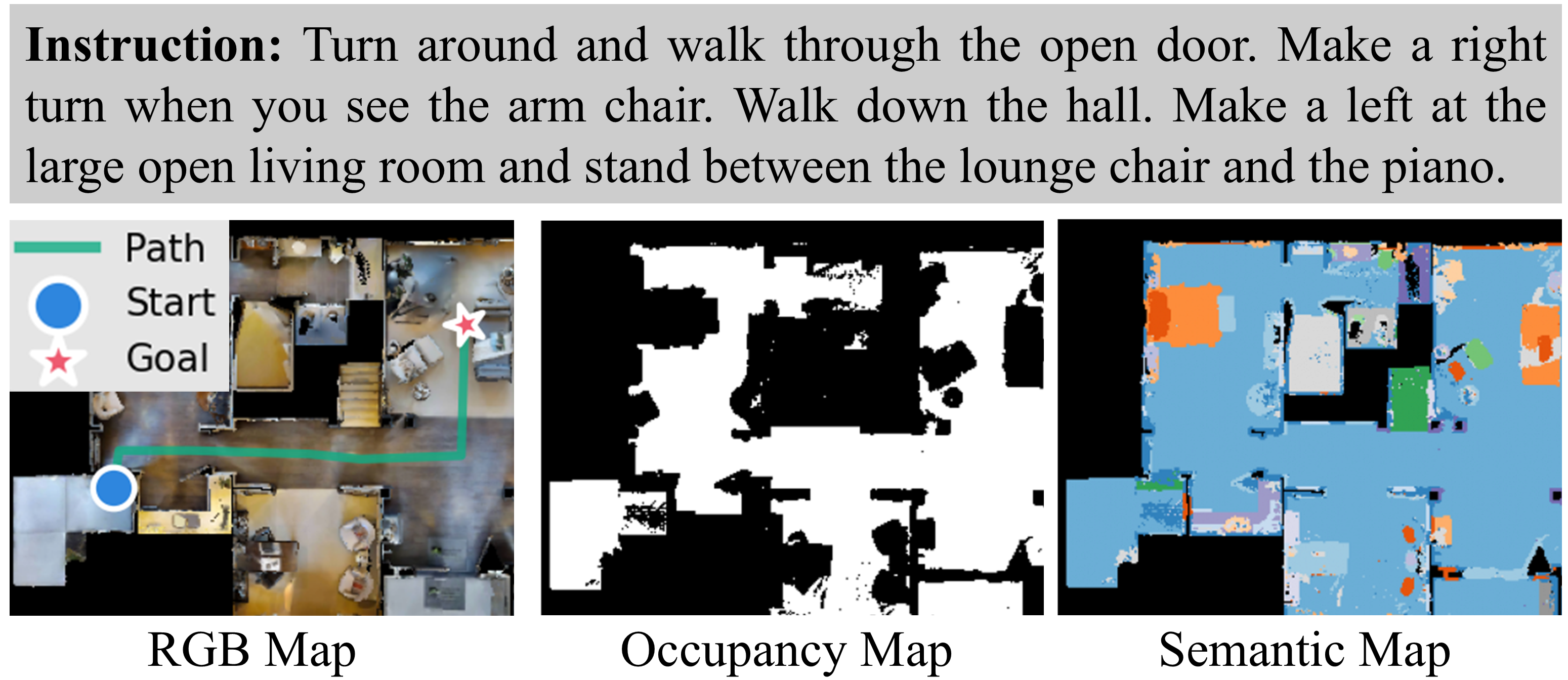}}%
\begin{minipage}[t]{0.56\linewidth}
\centering
\unhcopy5
\caption{Examples of multi-modal map inputs from R2R-TopDown. From left to right: RGB map, occupancy map (white=navigable, black=obstacle), semantic map (color-coded categories), and ground truth trajectory.}
\label{fig:dataset-examples}
\end{minipage}
\hfill
\begin{minipage}[t]{0.42\linewidth}
\centering
\includegraphics[width=\linewidth, height=\ht5]{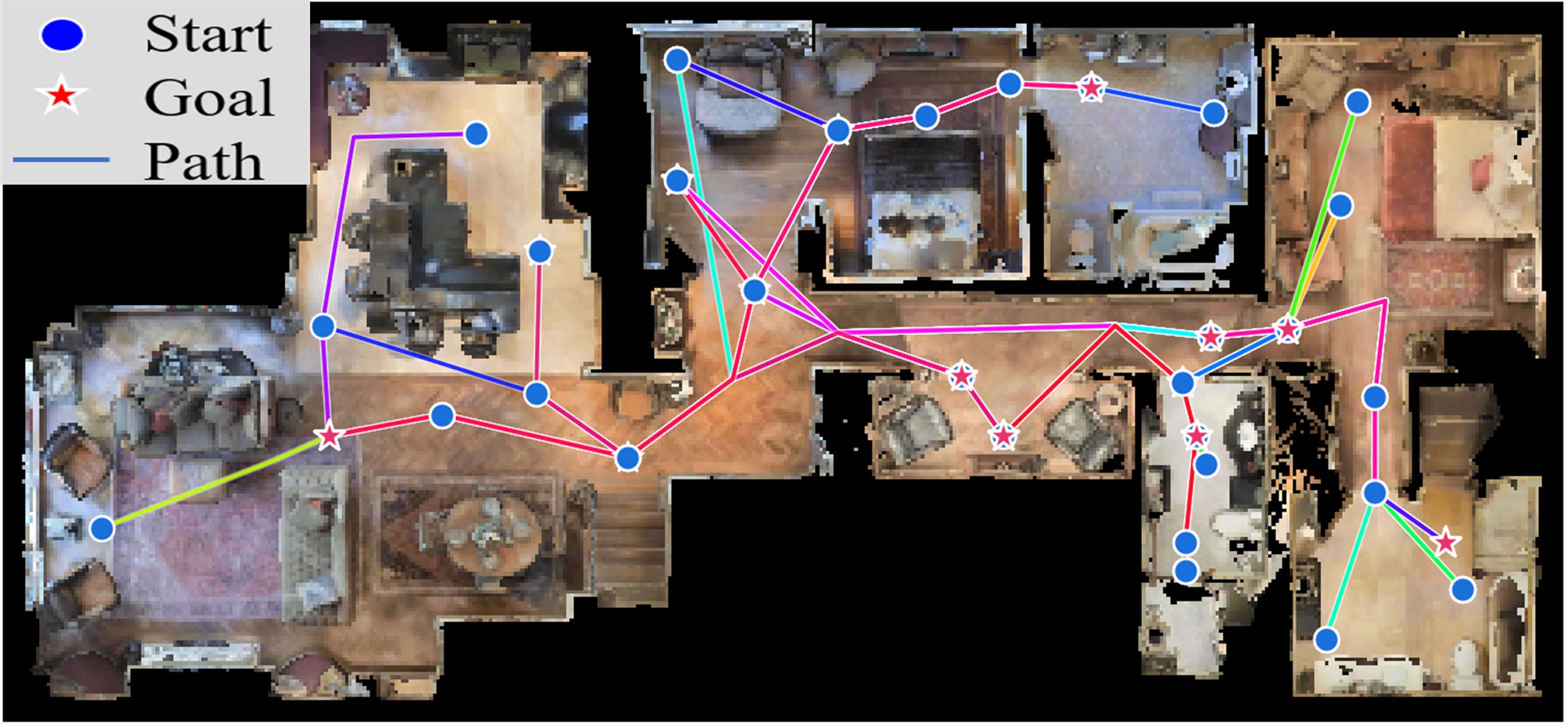}
\caption{Scene-level visualization of all training episodes. Each colored line is a distinct episode with start (blue circle) and goal (red star).}
\label{fig:scene-overview}
\end{minipage}
\end{figure}

\begin{wraptable}{r}{0.46\columnwidth}
\vspace{-1em}
\centering
\footnotesize
\setlength{\tabcolsep}{4pt}
\begin{tabular}{lccc}
\toprule
Metric & Train & Val Seen & Val Unseen \\
\midrule
Path Len.\ (m) & 9.58 & 9.92 & 9.83 \\
Instr.\ Len. & 26.5 & 27.3 & 26.8 \\
Total Samples & 6196 & 439 & 1003 \\
\bottomrule
\end{tabular}
\caption{Statistics of the R2R-TopDown dataset across training and validation splits.}
\label{tab:dataset-stats}
\vspace{-0.5em}
\end{wraptable}

\textbf{Dataset Statistics.}
The dataset comprises 6,196 training, 439 validation seen, and 1,003 validation unseen single-floor episodes across MP3D scenes. We exclude multi-floor trajectories from R2R-CE as they cannot be represented by a single top-down map. \Cref{tab:dataset-stats} summarizes key statistics across splits, and \Cref{fig:scene-overview} illustrates the spatial coverage, task diversity, and navigability of episodes across a representative scene.

\section{NavOne}
\label{sec:architecture}

\subsection{Task Definition and Preliminaries}
\label{sec:task-def}
We address the problem of VLN in continuous environments from a novel global perspective. Given a natural language instruction $L$, an RGB map $M_{rgb}$, an occupancy map $M_{occ}$, a semantic map $M_{sem}$, and the agent's initial state (position $p_0$ and rotation $r_0$), the goal is to generate a navigation path $\mathcal{P}$ that guides the agent to the goal location.

Let the input observation be defined as a multi-channel tensor $\mathbf{X} \in \mathbb{R}^{H \times W \times C}$, where $C$ is the total number of channels combining RGB, occupancy, and semantic information. The instruction $L$ is tokenized into a sequence of tokens $T = \{t_1, t_2, ..., t_N\}$. The agent's start position is a 2D coordinate $p_0 \in \mathbb{R}^2$ and a quaternion rotation $r_0 \in \mathbb{R}^4$.

\subsection{Method Overview}
\label{sec:method-overview}

\begin{figure*}[t]
\centering
\includegraphics[width=\linewidth]{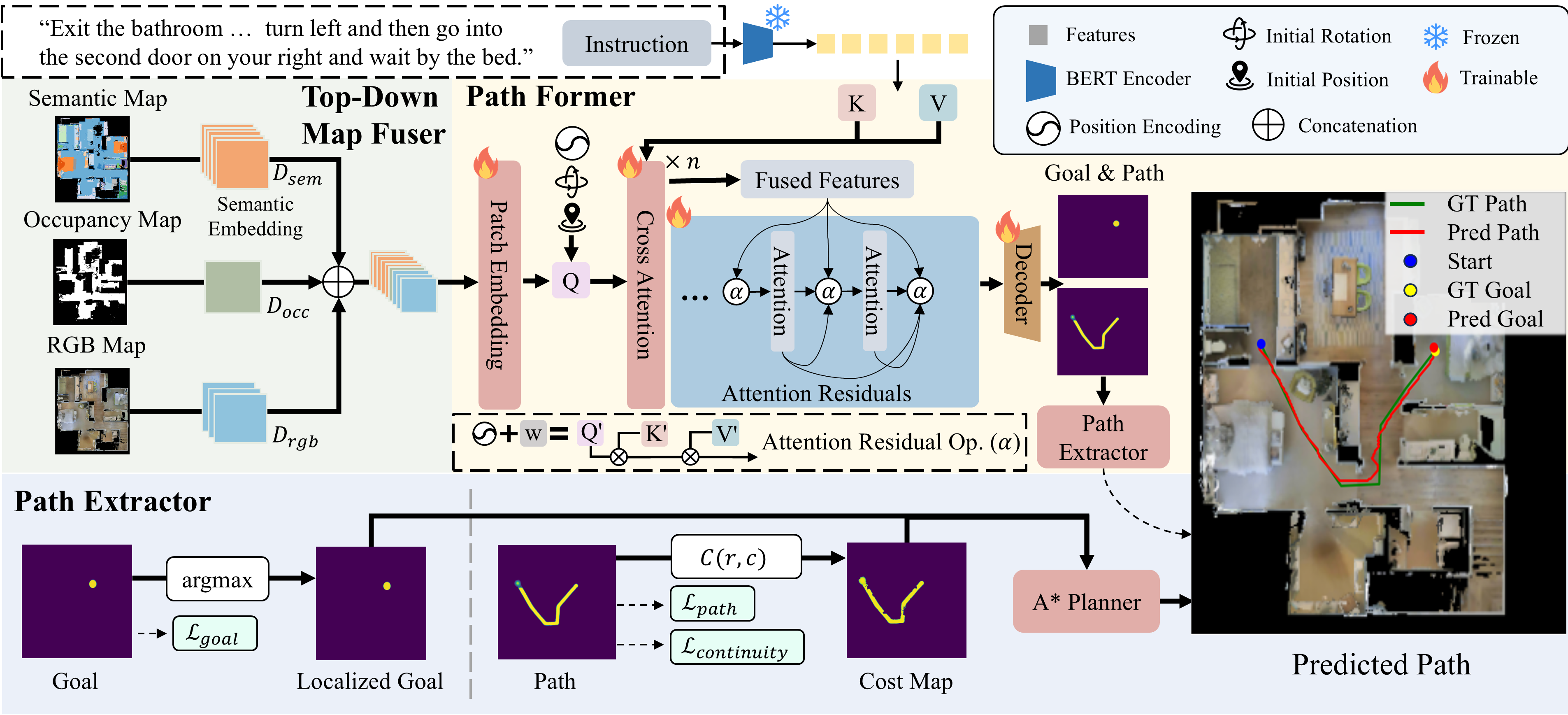}
\caption{Overview of our NavOne architecture. Multi-modal maps (RGB, occupancy, semantic) are concatenated and patch-embedded into visual tokens, while language instructions are encoded by BERT. Pose information (position and rotation) is injected via element-wise addition into visual tokens, which then attend to language features through stacked cross-attention layers. The resulting instruction-conditioned features are processed by a ViT encoder with Attention Residuals and decoded via progressive transposed convolutions to produce dense path and goal probability maps.}
\label{fig:architecture}
\end{figure*}

We propose NavOne, a unified end-to-end framework for the TD-VLN task that operates on global top-down maps. As illustrated in \Cref{fig:architecture}, NavOne comprises three components: 1)~a \textbf{Top-Down Map Fuser} that fuses multi-modal map inputs (RGB, occupancy, semantic) via channel-wise concatenation prior to patch embedding; 2)~a \textbf{Path Former} encoder-decoder that integrates language instructions, pose, and fused map features through stacked cross-attention and a ViT encoder equipped with Attention Residuals extended by spatial-aware depth queries for position-dependent feature mixing, then decodes to dense path and goal probability maps; and 3)~a \textbf{Path Extractor} that applies A* search over the predicted probability maps to compute the final executable navigation path. Unlike traditional egocentric VLN methods that require iterative action prediction at each step, NavOne generates the complete navigation path in a single forward pass, enabling efficient global spatial reasoning.

\subsection{Top-Down Map Fuser: Multi-Modal Map Representation}
The input to the model is a $C_{in}$-channel tensor:
\begin{equation}
\mathbf{X}_{input} = \text{Concat}(M_{rgb}, M_{occ}, \text{Embed}(M_{sem})),
\end{equation}
where $M_{rgb} \in \mathbb{R}^{H \times W \times 3}$ is the RGB map, $M_{occ} \in \mathbb{R}^{H \times W \times 1}$ is the occupancy map, and $\text{Embed}(M_{sem}) \in \mathbb{R}^{H \times W \times D_{sem}}$ is the embedded semantic map. We employ a learnable embedding layer to map discrete semantic categories into a $D_{sem}$-dimensional continuous feature space for efficient learning.

\subsection{Path Former: Encoder-Decoder Architecture}

\subsubsection{Multi-Modal Fusion}
We project heterogeneous inputs into a shared $D$-dimensional embedding space. The fused multi-modal map $\mathbf{X}_{input}$ is divided into $P \times P$ patches and linearly projected to obtain visual tokens $\mathbf{H}_{visual} \in \mathbb{R}^{N_v \times D}$, with learned patch positional embeddings $\{\mathbf{e}_n\}_{n=1}^{N_v}$ added token-wise, where $N_v = \frac{H}{P} \cdot \frac{W}{P}$ is the number of visual tokens. Language instructions are encoded by frozen BERT and projected to $\mathbf{H}_{text} \in \mathbb{R}^{N_t \times D}$, where $N_t$ is the number of text tokens. The agent's position $p_0 \in \mathbb{R}^2$ and rotation $r_0 \in \mathbb{R}^4$ are each linearly projected to $D$-dimensional vectors $\mathbf{h}_{pos}, \mathbf{h}_{rot} \in \mathbb{R}^D$.

We adopt a multi-modal fusion strategy. First, we anchor visual features with pose information via element-wise addition: $\mathbf{H}_{vp} = \mathbf{H}_{visual} + \mathbf{h}_{pos} + \mathbf{h}_{rot}$. Next, we employ $L_{cross}$ stacked cross-attention layers where pose-aware visual patches query language instructions:
\begin{equation}
\mathbf{H}_{fused}^{(l)} = \text{CrossAttention}^{(l)}(\mathbf{H}_{vp}^{(l-1)}, \mathbf{H}_{text}, \mathbf{H}_{text}),
\end{equation}
where $\mathbf{H}_{vp}^{(l-1)}$ serves as queries and $\mathbf{H}_{text}$ as keys and values. This progressive fusion allows visual tokens to retrieve task-relevant instruction information at multiple semantic levels. The final output $\mathbf{H}_{fused} = \mathbf{H}_{fused}^{(L_{cross})}$ captures instruction-conditioned spatial features.

\subsubsection{ViT Encoder with Attention Residuals}

We employ a ViT architecture as visual backbone, where the fused pose-aware and language-grounded features $\mathbf{H}_{fused}$ are processed through $L_{encoder}$ Transformer layers with Pre-LayerNorm~\cite{xiong2020layer} and $H$-head self-attention to extract global contextual features from the multi-modal maps with full receptive field.

However, standard transformer layers form a fixed chain in which each layer receives only its immediate predecessor's output, limiting direct access to representations from earlier depths. In top-down navigation, different spatial regions may benefit from different levels of abstraction. Moreover, a single global depth query imposes the same depth-mixing pattern across all spatial positions, motivating a per-token spatially conditioned formulation.

\textbf{Attention Residuals with Spatial-Aware Depth Query.}
We extend Attention Residuals~\cite{team2026attention} with \emph{spatial-aware depth queries} to enable position-dependent feature mixing across transformer layers. We adopt the \emph{Full Attention Residuals}, where every layer maintains a depth-wise attention mix over all previous layer outputs—as opposed to the Block Attention Residuals which groups layers and applies residual mixing only within blocks. Given our relatively shallow 12-layer ViT encoder, the Full Attention Residuals is computationally feasible and allows each layer to attend to the complete representation history without approximation. However, the original formulation uses a single global depth query shared across all spatial positions. We address this limitation by introducing a per-token \emph{spatial bias}:
\begin{equation}
    \mathbf{q}_l^{(n)} = \mathbf{w}_l + f_\theta(\mathbf{e}_n),
    \label{eq:spatial-depth-query}
\end{equation}
where each token's depth query is conditioned on its patch positional embedding $\mathbf{e}_n$ via a learned projection $f_\theta$. This introduces a position-conditioned modulation of depth mixing, allowing different spatial locations to express different preferences over representation depth. Both $\mathbf{w}_l$ and $f_\theta$ are zero-initialized, ensuring uniform depth mixing at the start of training for optimization stability.

\textbf{Decoder.}
To recover full spatial resolution from the ViT patch representation, we employ a lightweight progressive upsampling decoder. The Transformer output is reshaped to a spatial feature map and upsampled through multiple transposed convolution stages to resolution $H \times W$. A final $K \times K$ convolution produces the 2-channel output (path map and goal map).

\subsection{Path Extractor}
\label{sec:path-extraction}

The neural network outputs two probability maps: a path map $\mathbf{P}_{path} \in [0,1]^{H \times W}$ indicating the likelihood of each pixel belonging to the navigation path, and a goal map $\mathbf{P}_{goal} \in [0,1]^{H \times W}$ representing the predicted goal location distribution. We employ a modified A* search algorithm to extract the final discrete navigation path from these continuous predictions.

\textbf{Goal Localization.} The predicted goal position $g^* = (r_g, c_g)$ is obtained by taking the argmax over the goal probability map:
\begin{equation}
g^* = \arg\max_{(r,c)} \mathbf{P}_{goal}(r, c).
\end{equation}

\textbf{Cost Map Construction.} We construct a traversal cost map that combines the predicted path probabilities with occupancy constraints. For navigable regions, we invert the path probabilities to obtain traversal costs: pixels with high path probability receive low cost, encouraging the planner to follow the predicted trajectory. For non-navigable regions identified by the occupancy map, we assign a large penalty cost $C_{obs}$ to prevent traversal through obstacles. The complete cost formulation is:
\begin{equation}
C(r, c) = \begin{cases}
C_{obs} & \text{if } M_{occ}(r, c) = 0 \\
\frac{1}{\mathbf{P}_{path}(r, c) + \epsilon} & \text{otherwise}
\end{cases},
\end{equation}
where $\epsilon$ is a small constant to prevent division by zero.

\textbf{A* Search.} Given the start position $p_0$, predicted goal $g^*$, and cost map $C$, we apply A* search with 8-connectivity (including diagonal movements) to find the optimal path. The algorithm maintains the standard priority queue with $f$-scores:
\begin{equation}
f(n) = g(n) + h(n),
\end{equation}
where $g(n)$ denotes the accumulated cost from the start position to node $n$, and $h(n)$ is the Euclidean distance heuristic to the goal. When expanding to a neighbor node $n'$, the accumulated cost is updated as $g(n') = g(n) + C(n')$.

The search expands nodes in order of increasing $f$-score until reaching $g^*$, then reconstructs the path by backtracking through parent pointers. This formulation ensures that the extracted path follows regions with high predicted path probability, avoids obstacles indicated by the occupancy map, and reaches predicted goal location with shortest path.

\textbf{Failure Handling.} If A* cannot reach the predicted goal $g^*$, we snap the planning target to the nearest reachable free cell and rerun planning to recover an executable path.

\subsection{Optimization Objective}
\label{sec:optimization}
NavOne is trained with a multi-task objective that simultaneously predicts the path and goal location. Standard Binary Cross Entropy (BCE) loss is applied pixel-wise to both output maps. To further enforce smoothness and connectivity in the predicted path, a multi-component regularization term $\mathcal{L}_{continuity}$ is introduced, consisting of: $\mathcal{L}_{grad}$ to penalize large spatial gradients, $\mathcal{L}_{start}$ to enforce high path probabilities around the start position, and $\mathcal{L}_{erosion}$ utilizing morphological erosion to penalize fragmented predictions:
\begin{equation}
\mathcal{L}_{continuity} = \beta_1 \mathcal{L}_{grad} + \beta_2 \mathcal{L}_{start} + \beta_3 \mathcal{L}_{erosion}.
\end{equation}
This weighted combination promotes spatially smooth, connected paths anchored at the start position. Complete formulations are detailed in Appendix~\ref{appendix:loss-functions}. The final training objective is a weighted combination of path, goal, and continuity losses:
\begin{equation}
\mathcal{L}_{total} = \mathcal{L}_{path} + \alpha \mathcal{L}_{goal} + \lambda \mathcal{L}_{continuity},
\end{equation}
where $\mathcal{L}_{path}$ and $\mathcal{L}_{goal}$ are the standard BCE losses for their respective targets, and $\alpha$ and $\lambda$ control goal prediction and continuity regularization. Hyperparameter values are detailed in Appendix~\ref{appendix:implementation}.

\section{Experiments}
\label{sec:experiments}

\subsection{Datasets and Evaluation Metrics}

\begin{table*}[t]
\caption{Performance comparison with top-down map-based VLN methods on the R2R-TopDown dataset. All methods leverage global top-down maps for navigation. ``--'' indicates metrics not reported by the original paper. WS-MGMap, MapNav, and IPPD report results on the full R2R Val Unseen episodes; $\text{IPPD}^*$ denotes IPPD results on our filtered R2R-TopDown Val Unseen episodes, as the official paper only releases detailed results for the unseen split. WS-MGMap and MapNav numbers are taken directly from their original papers (full R2R Val Unseen); direct numerical comparison with NavOne (R2R-TopDown Val Unseen subset, 1003 episodes) should be interpreted with caution. Std.\ Attn: standard self-attention; AR-Full: Full Attention Residuals; SQ: Spatial-Aware Depth Query.}
\label{tab:main-results}
\centering
\begin{small}
\begin{tabular}{l cccc cccc}
\toprule
& \multicolumn{4}{c}{\textit{Val Seen}} & \multicolumn{4}{c}{\textit{Val Unseen}} \\
\cmidrule(lr){2-5} \cmidrule(lr){6-9}
Method & SR $\uparrow$ & SPL $\uparrow$ & TL $\downarrow$ & NE $\downarrow$ & SR $\uparrow$ & SPL $\uparrow$ & TL $\downarrow$ & NE $\downarrow$ \\
\midrule
WS-MGMap \cite{chen2022weakly} & 0.47 & 0.43 & 10.12 & 5.65 & 0.39 & 0.34 & 10.00 & 6.28 \\
MapNav \cite{zhang2025mapnav} & -- & -- & -- & -- & 0.40 & 0.37 & -- & \textbf{4.93} \\
IPPD \cite{wang2025instruction} & 0.57 & 0.54 & -- & -- & 0.45 & 0.42 & -- & -- \\
IPPD$^*$ \cite{wang2025instruction} & -- & -- & -- & -- & 0.37 & 0.31 & -- & -- \\
\textbf{NavOne (Std. Attn)} & \textbf{0.63} & \textbf{0.55} & 10.23 & 3.84 & 0.40 & 0.36 & 10.45 & 6.00 \\
\textbf{NavOne (AR-Full)} & 0.59 & 0.52 & \textbf{9.55} & \textbf{3.77} & 0.44 & 0.40 & 9.50 & 5.40 \\
\textbf{NavOne (AR-Full+SQ)} & 0.57 & 0.50 & 9.79 & 4.35 & \textbf{0.47} & \textbf{0.43} & \textbf{9.20} & 5.18 \\
\bottomrule
\end{tabular}
\end{small}
% \vspace{-2em}
\end{table*}

We evaluate our approach on the R2R-TopDown dataset and adopt standard VLN evaluation metrics. \emph{Success Rate (SR)} measures the percentage of episodes where the agent stops within 3 meters of the goal location. \emph{Success weighted by Path Length (SPL)} normalizes SR by the ratio of shortest path length to actual trajectory length, penalizing inefficient navigation. \emph{Trajectory Length (TL)} computes the average length of predicted navigation paths in meters. \emph{Navigation Error (NE)} measures the mean distance in meters between the agent's final position and the ground truth goal. Full implementation details including architecture specifications, training hyperparameters, and inference configurations are provided in Appendix~\ref{appendix:implementation}.

\subsection{Comparison with State-of-the-Art Methods}
\label{sec:comparison}

We compare NavOne against recent top-down map-based VLN methods: WS-MGMap~\cite{chen2022weakly}, which introduced weakly-supervised multi-granularity map learning; MapNav~\cite{zhang2025mapnav}, which leverages annotated semantic maps with a memory-based representation; and IPPD~\cite{wang2025instruction}, which applies diffusion-based path planning on indoor maps. We specifically compare against methods that operate within the global map paradigm rather than purely egocentric exploration models (e.g., ETPNav), ensuring a fair comparison where all methods benefit from the same pre-computed global spatial information. Results are summarized in \Cref{tab:main-results}.

On Val Seen, NavOne (Std.\ Attn) achieves the highest SR (0.63) and SPL (0.55), outperforming IPPD by +0.06 SR, while NavOne (AR-Full) yields the shortest trajectory (TL 9.55m) and the lowest navigation error (NE 3.77m). On Val Unseen, NavOne (AR-Full+SQ) achieves the highest SR (0.47) and SPL (0.43), surpassing WS-MGMap by +0.08 SR, MapNav by +0.07 SR, and IPPD by +0.02 SR. Notably, IPPD$^*$ (re-evaluated on our filtered episodes) yields a lower SR (0.37) than the original IPPD result (0.45), indicating that our R2R-TopDown Val Unseen subset is more challenging than the full validation set. MapNav reports the lowest navigation error (4.93m) among all compared methods; nonetheless, NavOne significantly outperforms it on the primary success metrics.

\subsection{Ablation on Architectural Components}
\label{sec:ablation}

The three NavOne variants in \Cref{tab:main-results} also serve as an ablation over key architectural components. Replacing AR-Full+SQ with standard self-attention (Std.\ Attn) improves Val Seen SR (0.63 vs.\ 0.57) but substantially hurts Val Unseen SR (0.40 vs.\ 0.47), suggesting weaker generalization to unseen map layouts. Adding Full Attention Residuals (AR-Full) recovers Val Unseen SR to 0.44 and narrows the seen--unseen gap from 23~pp to 15~pp, consistent with the benefit of mixing information from multiple network depths. Adding Spatial-Aware Depth Queries (AR-Full+SQ) further improves Val Unseen SR to 0.47 and reduces the gap to 10~pp, indicating that position-conditioned depth mixing is beneficial in our setting. To isolate the contribution of each map modality, we additionally ablate the input channels of the Top-Down Map Fuser; results are provided in Appendix~\ref{appendix:modality-ablation}.

\subsection{Computational Efficiency}
\label{sec:computational-efficiency}

We benchmark NavOne against two baselines on the same NVIDIA 4090D GPU: IPPD as a representative map-based planner, and ETPNav~\cite{an2024etpnav} as a representative sequential egocentric method.%
\clearpage

\begin{wraptable}{r}{0.42\columnwidth}
% \vspace{4em}
\centering
\begin{small}
\begin{tabular}{lcc}
\toprule
Method & Nav.\ Steps & Time (ms) \\
\midrule
ETPNav & 1$\sim$15 & 2970 \\
IPPD & N/A & 300 \\
\textbf{Ours} & 1 & \textbf{37} \\
\bottomrule
\end{tabular}
\end{small}
\caption{Avg.\ planning time per episode. N/A: IPPD uses a two-stage pipeline.}
\label{tab:inference-time}
% \vspace{-4em}
\end{wraptable}
All times are measured end-to-end from instruction input to final executable path output. This yields an $8\times$ speedup over IPPD (300ms) and a $\sim$80$\times$ reduction compared to ETPNav (2970ms), as shown in \Cref{tab:inference-time}. Note that the comparison with ETPNav reflects pure planning speed under the assumption of a pre-built map; ETPNav must build its representation online during exploration. The efficiency gain makes NavOne well-suited for rapid real-time re-planning in map-available settings.

\subsection{Qualitative Analysis}
\label{sec:qualitative}

\begin{wrapfigure}{r}{0.50\columnwidth}
\vspace{-1em}
\centering
\includegraphics[width=0.50\columnwidth]{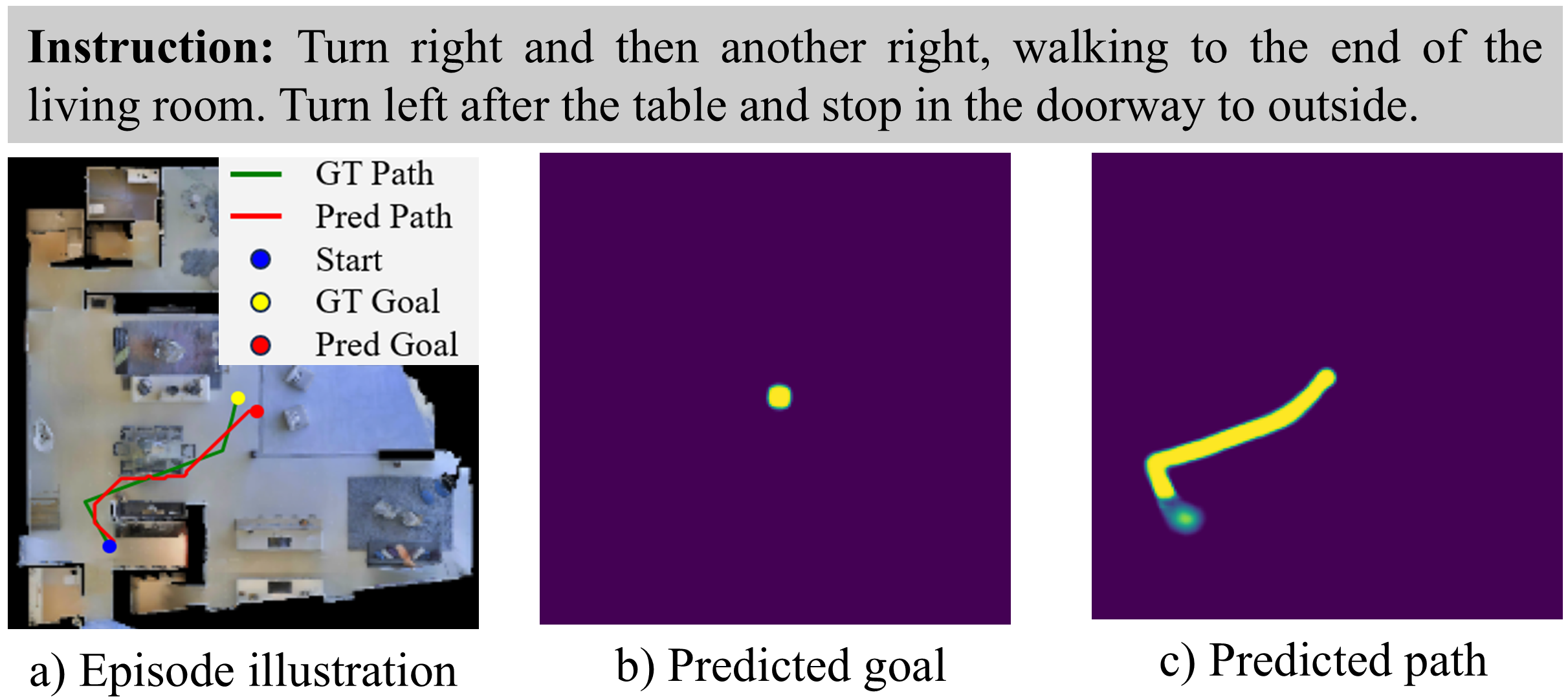}
\caption{Qualitative result: (a) predicted path (red) and ground truth (green) on the RGB map, (b) goal probability map, (c) path probability map.}
\label{fig:qualitative-results}
\vspace{-0.5em}
\end{wrapfigure}

We present a representative success case in \Cref{fig:qualitative-results}, where NavOne accurately grounds complex multi-step instructions into spatial predictions, correctly interpreting directional commands, spatial references, and landmark cues. The predicted goal map (b) shows a focused distribution at the correct doorway location, and the path map (c) exhibits high confidence along the navigable trajectory. The final A* path (red) closely matches the ground truth (green), validating that NavOne effectively bridges language understanding and geometric path planning. Additional qualitative examples are provided in Appendix~\ref{appendix:qualitative-results}.

\section{Discussion and Limitations}
\label{sec:limitations}

NavOne operates within a ``Map-then-Navigate'' paradigm, where multi-modal top-down maps are constructed once via standard SLAM pipelines (e.g., RTAB-Map~\cite{labbe2019rtab}) and semantic segmentation models (e.g., Mask2Former~\cite{cheng2022masked}), yielding several appealing properties: single-pass global path reasoning eliminates the iterative perception--action loop required by egocentric methods and, unlike discrete map-based methods, enables end-to-end learning from raw map inputs to continuous path outputs. Nevertheless, several limitations remain. First, maps constructed via SLAM may contain noise, and we evaluate NavOne's resilience to such combined map degradation (semantic noise, occupancy flips, and RGB corruption) in Appendix~\ref{appendix:robustness}. Second, real-world environments contain dynamic obstacles absent from the static pre-built map; future work will explore online map updates that incorporate real-time sensor observations to enable reactive replanning. Third, R2R instructions frequently reference fine-grained objects (e.g., photos, signs) absent from the 41-class semantic map vocabulary, limiting performance on unseen environments and motivating richer map representations.

\section{Conclusion}
\label{sec:conclusion}

We presented NavOne, a novel framework for Top-Down Vision-Language Navigation (TD-VLN) that formulates VLN as global planning on top-down maps by directly predicting interpretable path probability maps and goal probability maps from multi-modal spatial inputs and language. To support this paradigm, we introduce R2R-TopDown, a multi-modal benchmark derived from R2R-CE with RGB, occupancy, and semantic map representations. Results demonstrate state-of-the-art performance among top-down map-based methods with efficient single-pass inference. While TD-VLN assumes access to pre-computed maps, it complements exploration-based VLN and motivates future work on online mapping, multi-floor environments, and hybrid global–egocentric navigation systems.

\clearpage
\acknowledgments{This work was supported by Guangdong Provincial Natural Science Foundation for Outstanding Youth Team Project (No. 2024B1515040010), National Natural Science Foundation of China under Grant U23A20391, 62372188, and Guangdong Natural Science Foundation under Grant 2024A1515010100.}

% no \bibliographystyle is required, since the corl style is automatically used.
\bibliography{example_paper}

%%%%%%%%%%%%%%%%%%%%%%%%%%%%%%%%%%%%%%%%%%%%%%%%%%%%%%%%%%%%%%%%%%%%%%%%%%%%%%%
%%%%%%%%%%%%%%%%%%%%%%%%%%%%%%%%%%%%%%%%%%%%%%%%%%%%%%%%%%%%%%%%%%%%%%%%%%%%%%%
% APPENDIX
%%%%%%%%%%%%%%%%%%%%%%%%%%%%%%%%%%%%%%%%%%%%%%%%%%%%%%%%%%%%%%%%%%%%%%%%%%%%%%%
%%%%%%%%%%%%%%%%%%%%%%%%%%%%%%%%%%%%%%%%%%%%%%%%%%%%%%%%%%%%%%%%%%%%%%%%%%%%%%%
\newpage
\appendix

\section{R2R-TopDown Dataset Construction Details}
\label{appendix:dataset-construction}

This appendix provides detailed procedures for constructing the R2R-TopDown dataset from the original R2R-CE benchmark. We transform egocentric visual observations into multi-modal top-down map representations suitable for global path planning.

We construct the R2R-TopDown dataset through a multi-step process: 

\subsection{Environment Exploration and Map Construction}
\label{appendix:map-construction}

For each MP3D scene, we conduct a systematic pre-exploration pass to collect dense RGB-D observations and construct three aligned top-down maps. To ensure dataset accuracy, the entire map construction pipeline runs inside the Habitat simulator~\cite{savva2019habitat}, where ground-truth camera poses $\mathbf{T}_t \in SE(3)$ (full 3D rigid-body transformations used for point-cloud back-projection) are directly available from the simulator state rather than estimated from noisy sensor data. This eliminates SLAM drift and guarantees geometrically consistent maps, which is essential for producing reliable ground-truth path labels. We emphasize that this simulator-privileged access is used \emph{only} during offline dataset construction; in real-robot deployment, ground-truth poses are unavailable and a SLAM system must be used instead to estimate poses from onboard RGB-D streams, as detailed in \Cref{appendix:map-to-world}.

\textbf{Step 1: Frontier-Based Exploration.}
Starting from an arbitrary initial position, the agent navigates the scene using a frontier-based exploration strategy~\cite{yamauchi1997frontier}. The agent iteratively selects the nearest unexplored boundary (frontier) between known free space and unknown regions as its next navigation target, advancing until the coverage rate no longer improves. Throughout this process, the agent continuously captures RGB-D frames (RGB image and depth map) paired with the exact ground-truth pose $\mathbf{T}_t$ provided by the Habitat simulator at each step.

\textbf{Step 2: Pose Acquisition.}
We directly query the Habitat simulator for the ground-truth camera pose $\mathbf{T}_t \in SE(3)$ at each captured frame, yielding a globally consistent, drift-free pose sequence that serves as the geometric backbone for all subsequent map construction steps. Here $\mathbf{T}_t$ is the full 3D rigid-body transformation (3D translation + 3D rotation) required for RGB-D back-projection into world space; it is distinct from the agent's navigation input to NavOne, which uses only the 2D ground-plane position $p_0 \in \mathbb{R}^2$ and the quaternion heading $r_0 \in \mathbb{R}^4$. This simulator-privileged approach is specific to offline dataset construction. On a physical robot, ground-truth poses are unavailable; the same pipeline must instead rely on a SLAM system---such as RTAB-Map~\cite{labbe2019rtab}---to estimate globally consistent poses from the RGB-D stream via visual odometry, loop closure detection, and pose graph optimization.

\textbf{Step 3: RGB Map Generation.}
Using the ground-truth poses $\{\mathbf{T}_t\}$, each RGB-D frame is back-projected into the global 3D coordinate frame to form a colored point cloud $\mathcal{P}_{rgb}$. The top-down RGB map is obtained by orthographically projecting all points within the floor-level height band $[h_{min},\, h_{floor} + \Delta h]$ onto the horizontal plane, assigning each grid cell the mean color of its contained points:
\begin{equation}
M_{rgb}(x, z) = \frac{1}{|\mathcal{N}(x,z)|} \sum_{p \in \mathcal{N}(x,z)} c_p,
\end{equation}
where $\mathcal{N}(x,z)$ denotes the set of 3D points projected into cell $(x, z)$ and $c_p \in \mathbb{R}^3$ is the RGB color of point $p$.

\textbf{Step 4: Semantic Map Generation.}
A semantic segmentation model is applied to each RGB frame, producing per-pixel class labels from the MP3D 41-category vocabulary (see \Cref{tab:mp3d-categories}). The labels are back-projected to 3D using the same ground-truth depth and pose as in Step 3, forming a semantic point cloud $\mathcal{P}_{sem}$. For each grid cell, the final semantic label is determined by majority vote over all projected points:
\begin{equation}
M_{sem}(x,z) = \arg\max_{c} \sum_{p \in \mathcal{N}(x,z)} \mathbf{1}[l_p = c],
\end{equation}
where $l_p \in \{0,\dots,40\}$ is the semantic class of point $p$.

\textbf{Step 5: Occupancy Map Generation.}
The binary occupancy map is derived from the semantic map produced in Step 4. Pixels assigned the \textit{floor} category (MP3D category ID~2) are marked as navigable (free), while all other categories are treated as obstacles or background:
\begin{equation}
M_{occ}(x,z) = \begin{cases} 1 & \text{if } M_{sem}(x,z) = 2 \\ 0 & \text{otherwise} \end{cases},
\end{equation}
where $M_{occ}=1$ denotes a free (navigable) cell and $M_{occ}=0$ denotes an obstacle or background cell. This approach leverages the MP3D semantic annotation directly and requires no additional sensor processing.

\begin{table}[h]
\caption{The 41 semantic object categories in the Matterport3D (MP3D) dataset used for the semantic map $M_{sem}$. Category ID 0 (void) denotes unannotated or background pixels; IDs 1--40 are the 40 meaningful indoor object classes. Category ID 2 (floor) is also used to derive the binary occupancy map $M_{occ}$.}
\label{tab:mp3d-categories}
\centering
\footnotesize
\begin{tabular}{clclcl}
\toprule
ID & Category & ID & Category & ID & Category \\
\midrule
0  & void              & 14 & plant          & 28 & lighting \\
1  & wall              & 15 & sink           & 29 & beam \\
2  & floor             & 16 & stairs         & 30 & railing \\
3  & chair             & 17 & ceiling        & 31 & shelving \\
4  & door              & 18 & toilet         & 32 & blinds \\
5  & table             & 19 & stool          & 33 & gym\_equipment \\
6  & picture           & 20 & towel          & 34 & seating \\
7  & cabinet           & 21 & mirror         & 35 & board\_panel \\
8  & cushion           & 22 & tv\_monitor    & 36 & furniture \\
9  & window            & 23 & shower         & 37 & appliances \\
10 & sofa              & 24 & column         & 38 & clothes \\
11 & bed               & 25 & bathtub        & 39 & objects \\
12 & curtain           & 26 & counter        & 40 & misc \\
13 & chest\_of\_drawers & 27 & fireplace      &    & \\
\bottomrule
\end{tabular}
\end{table}

\textbf{Step 6: Coordinate Alignment and Cropping.}
All three maps share the global coordinate system provided by the Habitat simulator, so they are inherently co-registered with a uniform cell size. The maps are cropped to the bounding box of the explored region and resized to a unified resolution $H \times W$ for use in dataset construction.

\subsection{Real-Robot Map Construction with LingBot-Map}
\label{appendix:lingbot-map}

The dataset construction pipeline above relies on Habitat simulator ground-truth poses and depth maps, which are only available offline. For real-robot deployment, we use LingBot-Map~\cite{chen2026geometric}, a three-stage monocular mapping pipeline that reconstructs the same three-channel top-down map representation from an ordinary RGB video stream without any depth sensor or simulator access.

\textbf{Stage 1: Monocular Depth and Pose Estimation.}
A feed-forward visual reconstruction network processes the monocular RGB video frame-by-frame, jointly predicting per-frame metric depth maps and camera extrinsics $\{\mathbf{T}_t\} \subset SE(3)$. Depth predictions are confidence-filtered and back-projected into a global 3D point cloud using the estimated poses, analogous to the RGB-D back-projection in Steps 2--3 above, but operating without any depth sensor or ground-truth pose from a simulator.

\textbf{Stage 2: Semantic Segmentation via Mask2Former.}
Each RGB frame is passed through \textbf{Mask2Former}~\cite{cheng2022masked} with a Swin-L backbone pretrained on ADE20K (150 object categories). The ADE20K predictions are re-mapped to the MP3D-41 indoor vocabulary used during NavOne training via a hand-crafted category correspondence table, yielding per-pixel labels $l_p \in \{0,\dots,40\}$ consistent with the training-time semantic map $M_{sem}$. These labels are back-projected into the global point cloud together with the RGB colors estimated in Stage~1.

\textbf{Stage 3: Top-Down Map Assembly.}
The 3D point cloud is voxelised at a configurable resolution 1cm/pixel. A RANSAC-based floor-plane fitting step performs gravity alignment so that the floor lies at $Y=0$ in the voxel grid. Orthographic top-down projection then produces three co-registered output maps: \texttt{rgb} (mean color per cell), \texttt{sem} (MP3D-41 majority-vote class label per cell), and \texttt{occ} (binary traversability: $1$ = free/navigable, $0$ = obstacle). Map metadata---\texttt{meters\_per\_pixel} $s$ and world origin $(X_0, Z_0)$---is stored alongside the map and supports the same pixel-to-world coordinate mapping described in \Cref{appendix:map-to-world}.

\subsection{Map-to-World Coordinate Correspondence and Path Execution}
\label{appendix:map-to-world}

A key property of the top-down map representation is the bijective linear mapping between pixel coordinates and metric world coordinates, which directly resolves the question of how a predicted path on the map can be executed by a physical robot without additional egocentric alignment.

\textbf{Pixel-to-World Transformation.}
Let $(X_{origin}, Z_{origin})$ denote the world-frame position (in meters) corresponding to the top-left corner of the map, and let $s$ (m/pixel) denote the uniform cell size. For a pixel at row $r$ and column $c$, the corresponding world coordinates $(X, Z)$ are:
\begin{equation}
X = X_{origin} + c \cdot s, \qquad Z = Z_{origin} + r \cdot s.
\end{equation}
The inverse mapping from world coordinates to pixel coordinates is:
\begin{equation}
c = \left\lfloor \frac{X - X_{origin}}{s} \right\rfloor, \qquad r = \left\lfloor \frac{Z - Z_{origin}}{s} \right\rfloor.
\end{equation}
The parameters $(X_{origin},\, Z_{origin},\, s)$ are determined at map-construction time by the map-building pipeline and stored alongside the map, making every pixel coordinate unambiguously grounded in metric space.

\textbf{Execution Without Egocentric Alignment.}
During deployment, the same SLAM system used for map construction continuously estimates the robot's pose $\mathbf{T}_t \in SE(3)$ in the world frame. The robot's current 2D map position is obtained by extracting the XZ components of the translation part of $\mathbf{T}_t$ and applying the inverse pixel mapping above. NavOne's predicted path—an ordered sequence of pixel coordinates $\{(r_1, c_1), \dots, (r_K, c_K)\}$—is converted to metric waypoints $\{(X_1, Z_1), \dots, (X_K, Z_K)\}$ via the forward mapping. These waypoints are then passed to a low-level local controller (e.g., Dynamic Window Approach~\cite{fox2002dynamic}) as navigation targets. Critically, \emph{no explicit egocentric-to-map alignment is required}: the SLAM coordinate frame serves as the common reference for both the map and the robot's real-time localization, eliminating any misalignment between the global plan and the robot's perspective.

\textbf{Consistency with Dataset Encoding.}
The agent's initial state $(p_0, r_0)$ stored in every R2R-TopDown episode is expressed in the same world coordinate frame established during map construction. The original R2R-CE dataset stores agent poses natively as a 3D position and a unit quaternion $r_0 \in \mathbb{R}^4$; no rotation-format conversion is required. The 2D map position $p_0 \in \mathbb{R}^2$ is obtained by projecting the 3D position onto the ground plane (XZ plane) and converting to pixel coordinates via the inverse mapping above. This ensures that the start pose used as input to NavOne during both training and inference directly corresponds to the agent's localized pose in the map coordinate frame, maintaining full coordinate consistency across the pipeline.

\textbf{a) Top-Down Map Generation (Summary).} The above exploration pipeline produces three aligned 2D top-down maps per scene: an RGB map ($M_{rgb}$) providing a bird's-eye view of the environment, an occupancy map ($M_{occ}$) indicating navigable areas (white) versus obstacles (black), and a semantic map ($M_{sem}$) where each pixel encodes one of the 41 MP3D object categories.

\textbf{b) Trajectory Projection.} We compute the pixel coordinates of all waypoints on the top-down map and project the 3D navigation trajectory from R2R-CE onto the 2D map plane. The ground truth is represented as a sequence of 2D waypoints $\mathcal{W} = \{(x_1, y_1), \dots, (x_k, y_k)\}$ aligned with the top-down view.

\textbf{c) Agent State Encoding.} The dataset explicitly includes the agent's initial configuration: start coordinates ($p_0 \in \mathbb{R}^2$) and start orientation ($r_0 \in \mathbb{R}^4$ as a quaternion) within the global map frame.

\textbf{d) Single-Floor Filtering.} To ensure the feasibility of 2D map navigation, we filter the dataset to retain only episodes where the navigation trajectory remains on a single floor. For MP3D scenes where each environment consists of multiple floors with separate top-down views, we manually exclude trajectories that span multiple floors or floors with poor reconstruction quality.

After applying these filters, the R2R-TopDown dataset contains 6,196 training trajectories, 439 validation seen (Val Seen), and 1,003 validation unseen (Val Unseen) episodes. While smaller than the original R2R-CE due to the single-floor constraint, the filtered dataset ensures that all episodes can be reliably represented and solved using 2D top-down map navigation.

\section{Loss Function Components}
\label{appendix:loss-functions}

This appendix provides the full mathematical formulations for the components of the Path Continuity Loss $\mathcal{L}_{continuity}$ discussed in Section~\ref{sec:optimization}.

First, we apply Sobel operators to compute spatial gradients and penalize large gradient magnitudes via 
\begin{equation}
\mathcal{L}_{grad} = \mathbb{E}_{x,y} \left[ \sqrt{(\partial \mathbf{P}_{path}/\partial x)^2 + (\partial \mathbf{P}_{path}/\partial y)^2} \right],
\end{equation}
encouraging smooth transitions in path probabilities. 

Second, we enforce high path probabilities around the start position $p_0$ using a mean squared error loss with linearly decaying targets: for pixels within radius $r_s$, the target probability is $p_{target} = \max(0, 1 - d/r_s)$ where $d$ is the Euclidean distance, and the loss is computed as:
\begin{equation}
\mathcal{L}_{start} = \mathbb{E}_{d < r_s} \left[ (\mathbf{P}_{path}(x,y) - p_{target})^2 \right],
\end{equation}
ensuring the path originates from the correct location with high confidence. 

Third, we apply morphological erosion ($k \times k$ averaging kernel) to penalize fragmented predictions, since continuous paths should remain largely intact after erosion while disconnected segments degrade significantly: 
\begin{equation}
\mathcal{L}_{erosion} = \mathbb{E} [ \text{ReLU}(\mathbf{P}_{path} - \text{Erode}_k(\mathbf{P}_{path}) - \tau_e) ],
\end{equation}
where $\tau_e$ is the erosion tolerance threshold.

\section{Implementation Details}
\label{appendix:implementation}

\subsection{Architecture \& Inference}
We employ ViT-B/16 as the visual backbone with patch size $P=16$, hidden dimension $D=768$, $L_{encoder}=12$ transformer layers with Pre-LayerNorm~\cite{xiong2020layer}, $H=12$ attention heads, and attention dropout rate $0.1$. The input consists of $C_{in}=12$ channels constructed by concatenating RGB maps (3 channels), occupancy map (1 channel), and semantically embedded maps with $D_{sem}=8$ channels. Language instructions are encoded using BERT-base-uncased, which is kept frozen during training to leverage its pretrained linguistic knowledge.

For multi-modal fusion, we use $L_{cross}=12$ stacked cross-attention layers to progressively integrate language instructions with pose-aware visual features before feeding them into the ViT encoder.

\textbf{Attention Residuals Configuration.} We employ the Full Attention Residuals variant~\cite{team2026attention} where every layer's input is a depth-wise softmax-weighted mix of all previous layer outputs. The spatial-aware depth query is enabled: each token's depth query is $\mathbf{q}_l^{(n)} = \mathbf{w}_l + f_\theta(\mathbf{e}_n)$, where $f_\theta$ is a zero-initialized linear projection. All per-layer depth queries $\mathbf{w}_l$ are also zero-initialized, ensuring uniform mixing at training start for stability.

The decoder employs a progressive upsampling architecture to recover spatial resolution from the compressed ViT encoder output. Starting from the encoded features at grid resolution $16\times16$ with $D=768$ channels, the decoder applies a series of transposed convolution stages that simultaneously upsample spatial dimensions and reduce channel dimensions. Specifically, the channel dimensionality progressively decreases through $768\rightarrow256\rightarrow128\rightarrow64\rightarrow32$, while spatial resolution is doubled at each stage via stride-2 transposed convolutions with $2\times2$ kernels until reaching the target size of $H \times W = 256 \times 256$. Each upsampling stage is followed by batch normalization and ReLU activation. A final $3\times3$ convolutional layer projects the 32-dimensional features to 2 output channels corresponding to the path probability map and goal probability map.

At inference time, the A* search algorithm operates on the predicted probability maps with the following hyperparameters: $\epsilon=0.01$ for numerical stability in the cost map construction, obstacle penalty cost $C_{obs}=1000$ to strongly discourage traversal through non-navigable regions, and 8-connectivity neighborhood structure enabling diagonal movements with Euclidean distance as the heuristic function.

\subsection{Training \& Optimization}
We train the model for 400 epochs with batch size 30 using the AdamW optimizer. The learning rate is set to $\text{lr}=10^{-4}$ with weight decay $0.05$. We employ a cosine annealing learning rate schedule with minimum learning rate $\eta_{\min}=10^{-6}$ and a 30-epoch linear warmup phase at the beginning of training to stabilize early optimization. Gradient clipping with max norm 1.0 is applied throughout training.

Data augmentation is applied during training to improve generalization: random rotation (uniform sampling from $[-15^\circ, 15^\circ]$), random translation (maximum 15\% of image dimensions in both horizontal and vertical directions), and color jitter (brightness/contrast/saturation adjusted within $\pm30\%$, hue within $\pm10\%$). Training takes approximately 14 hours on a single NVIDIA 4090D GPU (24GB VRAM).

For the multi-task loss function, we set goal prediction weight $\alpha=5.0$ to emphasize accurate goal localization, and continuity regularization weight $\lambda=0.1$ to enforce smooth path predictions. The path continuity loss uses the following hyperparameters: start position enforcement radius $r_s=8$ pixels, morphological erosion kernel size $k=3\times3$, erosion tolerance threshold $\tau_e=0.3$, and continuity component weights $(\beta_1, \beta_2, \beta_3)=(0.2, 0.5, 0.3)$ balancing gradient smoothness, start position anchoring, and erosion-based connectivity penalties respectively.

\section{Feature Orthogonality Analysis}
\label{sec:orthogonality-analysis}

NavOne injects pose information into visual tokens via element-wise addition ($\mathbf{H}_{vp} = \mathbf{H}_{visual} + \mathbf{h}_{pos} + \mathbf{h}_{rot}$). We analyze the orthogonality of learned feature representations to validate why this parameter-free operation is sufficient. When different modalities encode semantically distinct information (visual appearance, instructions, spatial position, and orientation), their learned representations tend to occupy near-orthogonal subspaces in the high-dimensional feature space. Under this orthogonality assumption, element-wise addition preserves each modality's information without interference: projecting the fused representation onto any modality's subspace recovers approximately the original signal, analogous to lossless superposition of orthogonal basis vectors.

We extract features from validation samples and compute pairwise cosine similarities and angular distances between the four modalities to validate this hypothesis.

\begin{table}[h]
\centering
\caption{Angular distances (in degrees) between modality pairs in the learned feature space. Values closer to 90° indicate better orthogonality. Diagonal entries are 0° as each modality is perfectly aligned with itself.}
\label{tab:angle-matrix}
\begin{small}
\begin{tabular}{lcccc}
\toprule
& Visual & Instruction & Initial Rotation & Initial Position \\
\midrule
Visual Input     & 0.0   & 88.6 & 98.1 & 88.1 \\
Instruction      & 88.6  & 0.0  & 92.0 & 86.6 \\
Initial Rotation & 98.1  & 92.0 & 0.0  & 96.4 \\
Initial Position & 88.1  & 86.6 & 96.4 & 0.0  \\
\bottomrule
\end{tabular}
\end{small}
\end{table}

As shown in \Cref{tab:angle-matrix}, most feature pairs exhibit angles close to 90°, indicating strong orthogonality. The vision encoder and language encoder learn highly complementary representations (visual--instruction: 88.6°), with text encoding high-level semantic goals and visual features capturing spatial structure. Linguistic features remain nearly orthogonal to pose embeddings (instruction--rotation: 92.0°, instruction--position: 86.6°). The visual--rotation (98.1°), visual--position (88.1°), and rotation--position (96.4°) pairs also demonstrate substantial orthogonality. The near-orthogonality confirms that the network naturally learns to encode each modality in a distinct subspace, validating that element-wise addition is a sufficient operation for pose injection: when features are orthogonal, linear superposition preserves each modality's information without destructive interference.

\section{Additional Qualitative Results}
\label{appendix:qualitative-results}

To provide further insight into our model's navigation capabilities, we present additional qualitative examples showcasing successful path predictions across diverse instructions and spatial configurations. For all examples, we visualize: (Left) path predictions with predicted path in red and ground truth in green overlaid on the RGB map; (Middle) predicted goal probability map; (Right) predicted path probability map.

\begin{figure}[h]
\centering
\begin{minipage}[t]{0.48\linewidth}
\centering
\includegraphics[width=\linewidth]{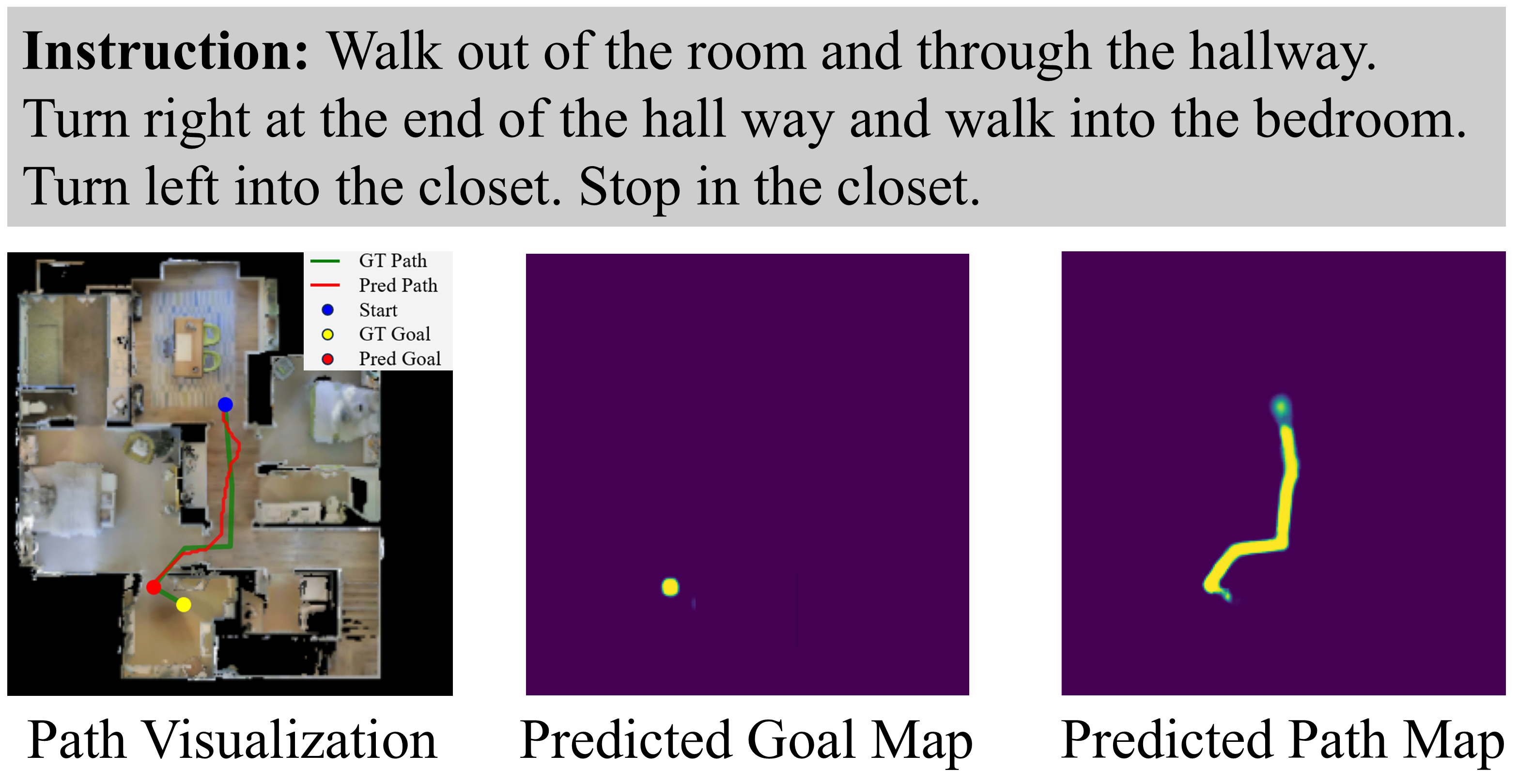}
\caption{Multi-room navigation example.}
\label{fig:appendix-qual-1}
\end{minipage}
\hfill
\begin{minipage}[t]{0.48\linewidth}
\centering
\includegraphics[width=\linewidth]{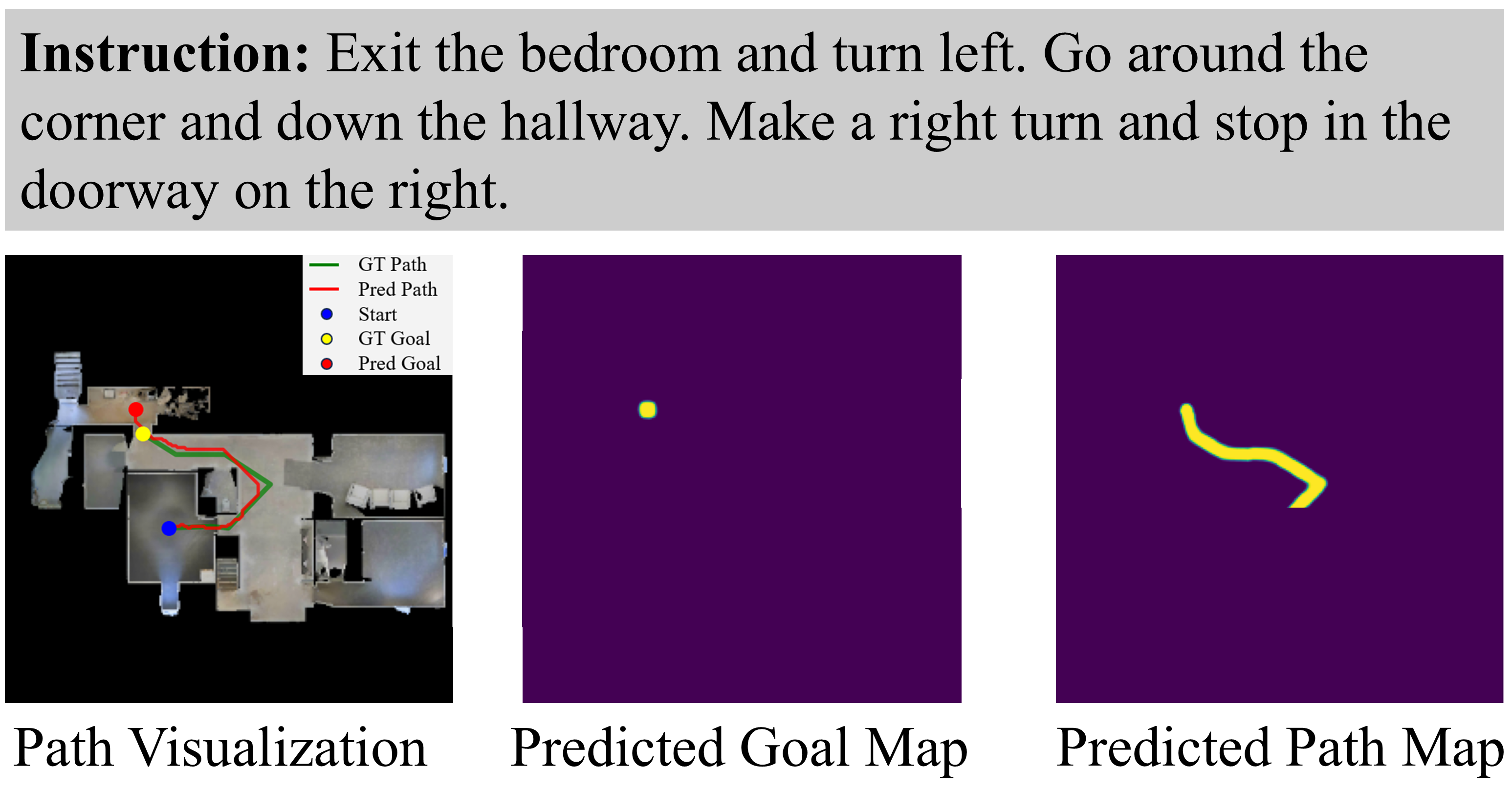}
\caption{Long-corridor navigation example.}
\label{fig:appendix-qual-2}
\end{minipage}
\end{figure}

\Cref{fig:appendix-qual-1} illustrates a complex multi-room scenario where the instruction requires sequential waypoints: exiting the initial room, traversing a hallway, turning right into a bedroom, and finally turning left into a closet. The model successfully generates accurate predictions with strong goal activation at the closet location and high-confidence path probability along the entire trajectory, correctly capturing all turns and room transitions.

\Cref{fig:appendix-qual-2} demonstrates navigation through extended corridors with accurate goal and path predictions, maintaining high confidence along the navigable route while correctly avoiding non-traversable regions.

\begin{figure}[h]
\centering
\begin{minipage}[t]{0.48\linewidth}
\centering
\includegraphics[width=\linewidth]{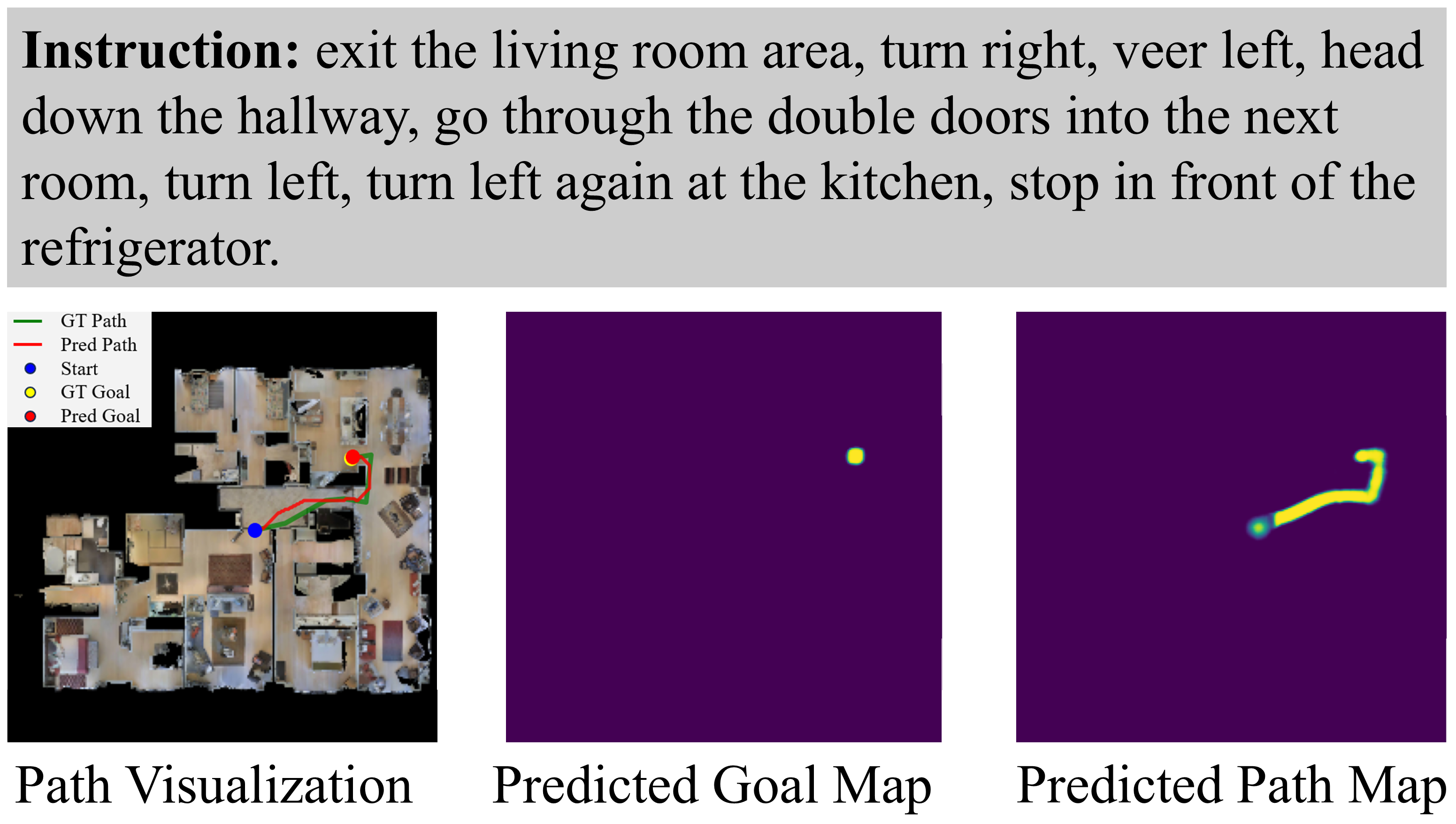}
\caption{Kitchen navigation example.}
\label{fig:appendix-qual-3}
\end{minipage}
\hfill
\begin{minipage}[t]{0.48\linewidth}
\centering
\includegraphics[width=\linewidth]{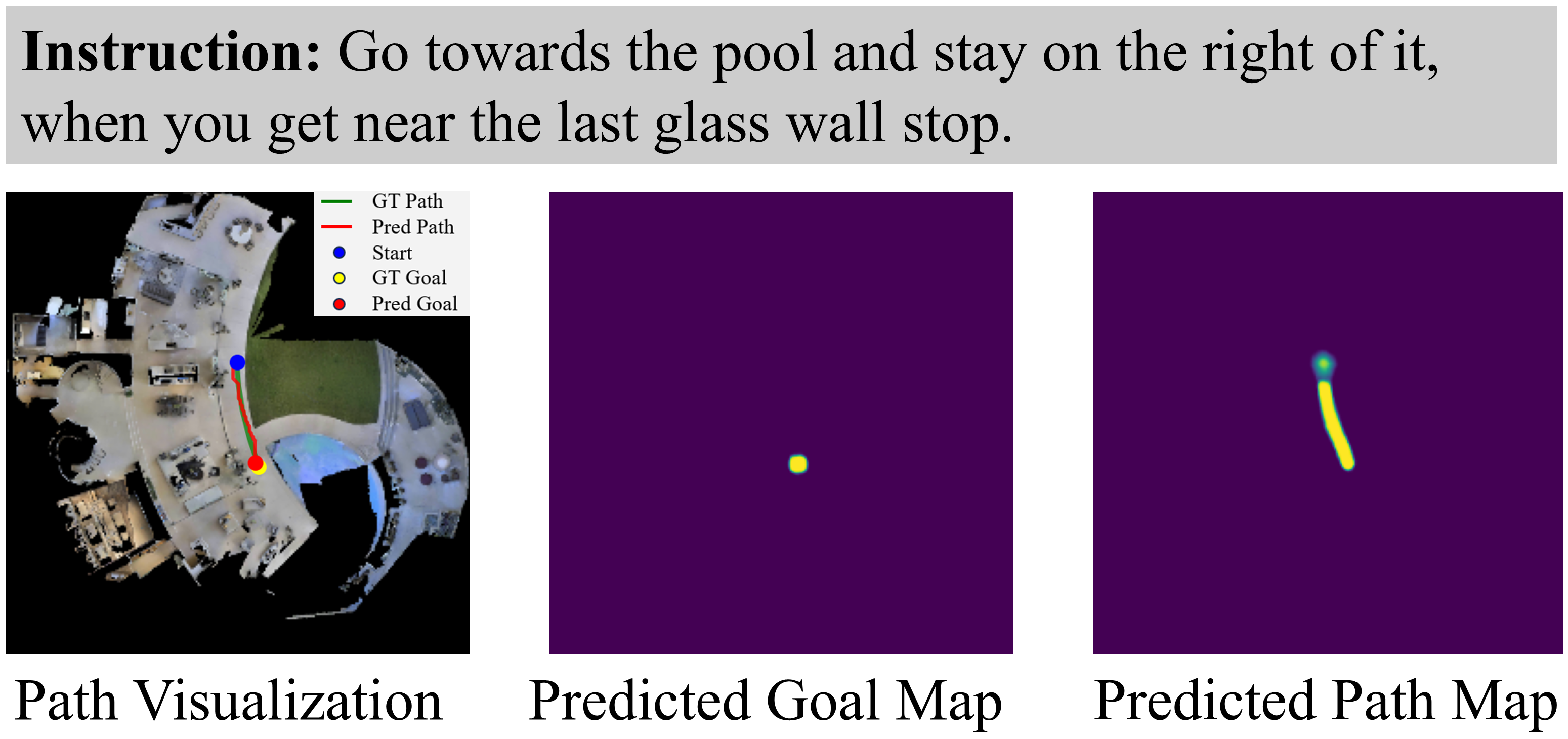}
\caption{Outdoor pool area navigation example.}
\label{fig:appendix-qual-4}
\end{minipage}
\end{figure}

\Cref{fig:appendix-qual-3} demonstrates a highly complex task with numerous sequential actions: exiting the living room, turning right, veering left, heading down the hallway, going through double doors, turning left twice at the kitchen, and stopping in front of the refrigerator. Despite the intricate instruction, the model produces sharp goal activation at the refrigerator and maintains strong path confidence throughout the lengthy trajectory.

\Cref{fig:appendix-qual-4} showcases an outdoor pool area scenario requiring spatial relationship understanding (``stay on the right side''). The model correctly follows the right side of the pool and accurately identifies the stopping position near the glass wall, demonstrating its ability to handle relative positioning constraints.

These examples collectively validate our model's capabilities across diverse scenarios: multi-room navigation with sequential waypoints, long corridors, complex instruction sequences with numerous turns, and outdoor environments with spatial constraints. The consistently accurate goal localization and high-confidence path predictions across both indoor and outdoor settings confirm the robustness and generalization capability of our global path planning approach.

\textbf{Real-Robot Map Generalization.}
To validate that NavOne generalizes beyond simulator-constructed maps, we run inference directly on top-down maps built by the LingBot-Map pipeline~(\Cref{appendix:lingbot-map}) from monocular RGB video of a real indoor corridor, without any fine-tuning.

\begin{figure}[h]
\centering
\begin{minipage}[t]{0.48\linewidth}
\centering
\includegraphics[width=\linewidth]{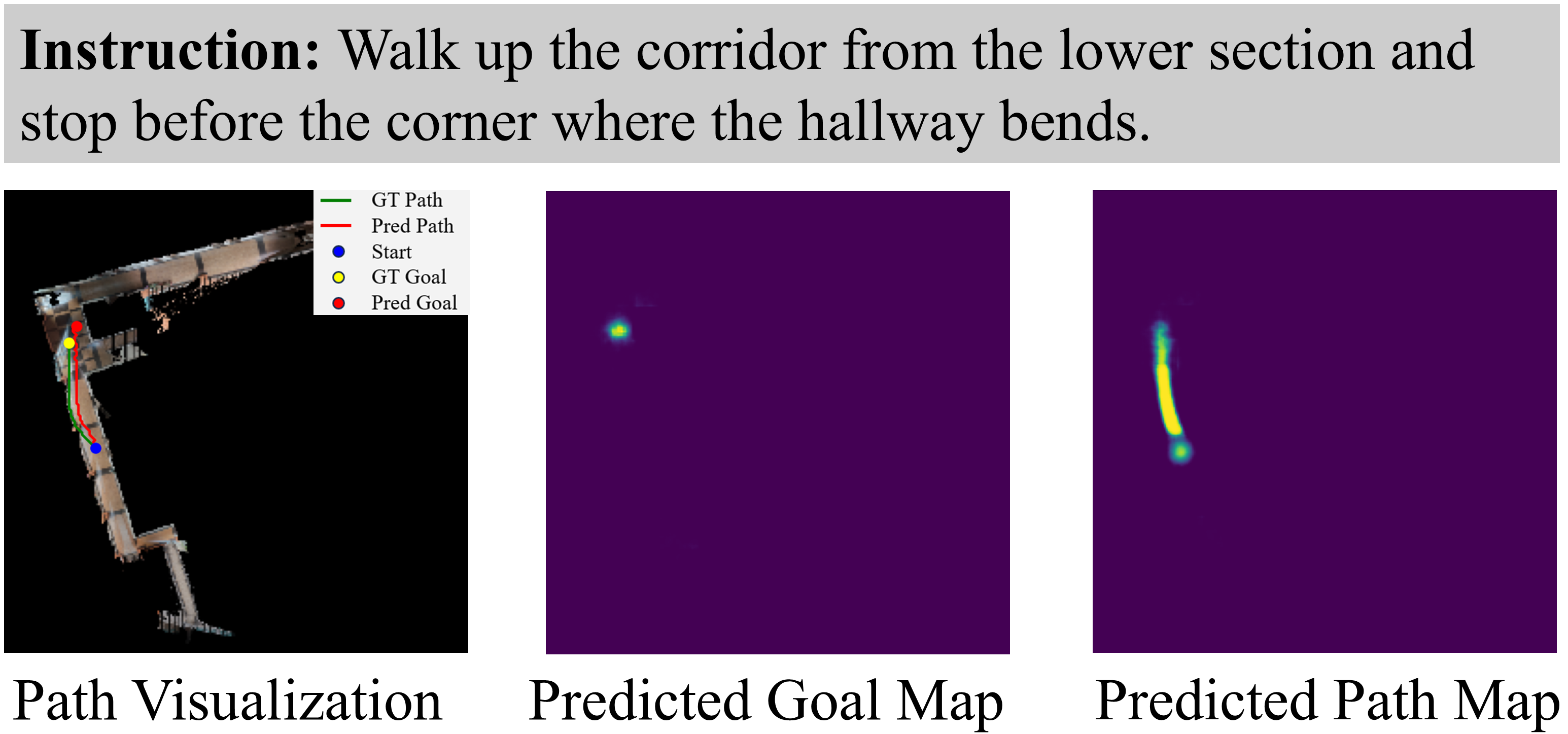}
\caption{Real-robot corridor navigation example~1.}
\label{fig:appendix-real-exp1}
\end{minipage}
\hfill
\begin{minipage}[t]{0.48\linewidth}
\centering
\includegraphics[width=\linewidth]{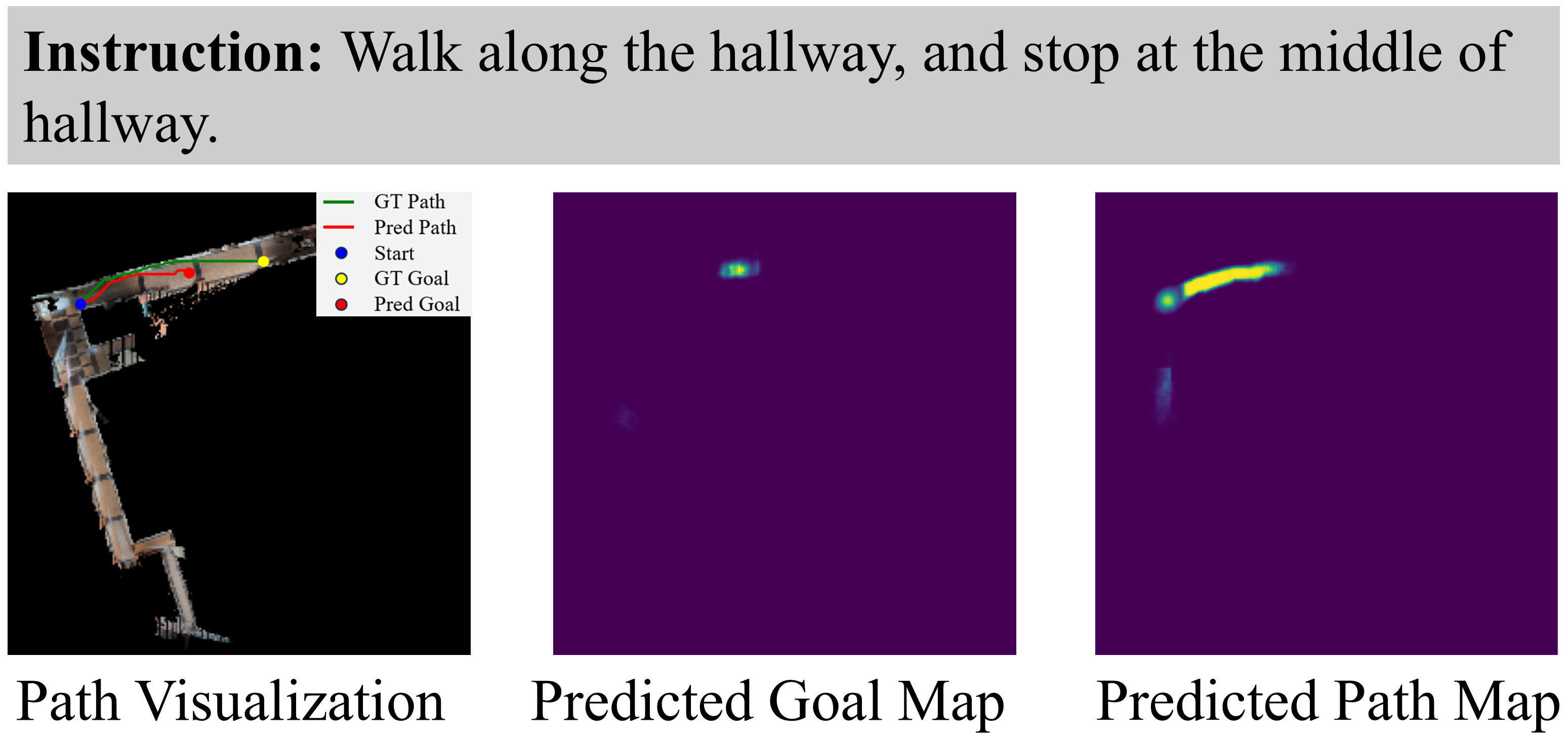}
\caption{Real-robot corridor navigation example~2.}
\label{fig:appendix-real-exp2}
\end{minipage}
\end{figure}

\Cref{fig:appendix-real-exp1,fig:appendix-real-exp2} show two representative corridor navigation cases. In both examples, NavOne produces accurate goal probability peaks at the correct destinations and path probability maps that trace the navigable corridor, closely matching the ground-truth A* paths. These results demonstrate that the three-modal input representation (RGB, semantic, occupancy) learned on simulator-generated R2R-TopDown data transfers directly to maps produced by a real-world monocular mapping pipeline, without any domain adaptation.

\section{Robustness to Map Degradation}
\label{appendix:robustness}

\begin{table}[h]
\caption{Robustness to combined map degradation on Val Unseen. All three noise types are applied simultaneously. \textbf{No Aug}: standard training on clean maps. \textbf{Noise Aug}: trained with 10\% combined noise augmentation. SR: Success Rate; SPL: Success weighted by Path Length.}
\label{tab:robustness}
\centering
\footnotesize
\begin{tabular}{lcccccc}
\toprule
    & & \multicolumn{2}{c}{\textbf{No Aug}} & \multicolumn{2}{c}{\textbf{Noise Aug}} \\
\cmidrule(lr){3-4} \cmidrule(lr){5-6}
Noise Level & & SR $\uparrow$ & SPL $\uparrow$ & SR $\uparrow$ & SPL $\uparrow$ \\
\midrule
None (clean)  & & 0.471 & 0.439 & 0.483 & 0.450 \\
\midrule
Combined 5\%  & & 0.448 & 0.412 & 0.474 & 0.442 \\
Combined 10\% & & 0.436 & 0.399 & 0.469 & 0.434 \\
Combined 20\% & & 0.368 & 0.319 & 0.442 & 0.404 \\
Combined 30\% & & 0.297 & 0.249 & 0.420 & 0.364 \\
\bottomrule
\end{tabular}
\end{table}

A key concern for practical deployment is whether NavOne degrades gracefully when maps are imperfect---as is inevitable with real-world SLAM and semantic segmentation pipelines. We evaluate robustness under a challenging \textbf{combined noise} protocol that simultaneously applies three corruption types to the input map tensor:
(i)~\textbf{Semantic Random}: replace a fraction of semantic pixels with random class labels, simulating misclassification errors from the segmentation model;
(ii)~\textbf{Occupancy Flip}: randomly flip occupancy cell values between free and occupied, simulating SLAM mapping errors;
(iii)~\textbf{RGB Gaussian}: add zero-mean Gaussian noise to RGB channels, simulating sensor or compression artifacts.
The noise level parameter controls the corruption fraction (or $\sigma$) uniformly across all three types. We compare two training strategies: the default NavOne (AR-Full+SQ) model trained on clean maps (\textbf{No Aug}) and the same model trained with 10\% combined noise augmentation (\textbf{Noise Aug}).

Two observations emerge from \Cref{tab:robustness}. First, \textbf{the standard model degrades substantially under severe combined noise}: SR falls from 0.471 to 0.297 at the 30\% level (17.4~pp drop), reflecting the compounding effect of simultaneous corruptions across all map modalities. Second, \textbf{noise augmentation training dramatically improves robustness with no clean-map penalty}: the noise-augmented model not only retains higher SR under all corruption levels (0.420 vs.\ 0.297 at 30\%, a 12.3~pp gap), but also achieves slightly better clean performance (0.483 vs.\ 0.471 SR), suggesting that exposure to diverse map artifacts acts as a regularizer. At the moderate 20\% level, the augmented model closes the gap by 7.4~pp (0.442 vs.\ 0.368), demonstrating that a lightweight augmentation strategy applied during training is sufficient to substantially harden NavOne against realistic multi-modal map degradation.

\section{Input Modality Ablation}
\label{appendix:modality-ablation}

\begin{table}[h]
\caption{Ablation on input modalities. Metrics are reported on the R2R-TopDown Val Seen and Val Unseen splits. Navigation Error (NE) and Trajectory Length (TL) are in meters, converted from the internal evaluation unit by dividing by 20 (same as \Cref{tab:main-results}).}
\label{tab:modality-ablation}
\centering
\footnotesize
\begin{tabular}{l cccc cccc}
\toprule
& \multicolumn{4}{c}{\textit{Val Seen}} & \multicolumn{4}{c}{\textit{Val Unseen}} \\
\cmidrule(lr){2-5} \cmidrule(lr){6-9}
Method & SR $\uparrow$ & SPL $\uparrow$ & TL $\downarrow$ & NE $\downarrow$ & SR $\uparrow$ & SPL $\uparrow$ & TL $\downarrow$ & NE $\downarrow$ \\
\midrule
RGB only & 0.36 & 0.31 & \textbf{9.64} & 6.33 & 0.20 & 0.18 & 9.26 & 7.02 \\
Semantic only & 0.53 & 0.47 & 10.29 & 4.54 & 0.43 & 0.39 & 10.03 & 5.48 \\
\textbf{NavOne (AR-Full+SQ)} & \textbf{0.57} & \textbf{0.50} & 9.79 & \textbf{4.35} & \textbf{0.47} & \textbf{0.43} & \textbf{9.20} & \textbf{5.18} \\
\bottomrule
\end{tabular}
\end{table}

To quantify the contribution of each map modality, we train NavOne (AR-Full+SQ) with a single input channelization while keeping the same architecture, loss, and training recipe as the full model: we either feed only the RGB top-down map (RGB only), or only the embedded semantic map (semantic only). We do \textbf{not} report an \emph{occupancy-only} model: the occupancy map is a binary free-space layer without room categories, object identities, or appearance cues, so the network cannot match natural-language references (e.g., \emph{``turn left at the kitchen''} or \emph{``stop by the table''}) to specific regions of the map; optimization collapses to unconstrained A* in free space, which is ill-suited to instruction-following. Therefore occupancy is included as a necessary geometric prior in the full three-modal setting rather than as a stand-alone input.

Results in \Cref{tab:modality-ablation} show that a single map layer is far weaker than the full three-modal Top-Down Map Fuser. \textbf{RGB only} reaches only 0.36~SR on Val Seen and drops sharply to 0.20~SR on Val Unseen, indicating that photometric appearance alone is insufficient to ground long-horizon, object-centric instructions in global layout, with the large seen-unseen gap (16~pp) reflecting poor generalization across environments. \textbf{Semantic only} attains 0.53/0.43~SR on Val Seen/Unseen, substantially outperforming RGB only, because discrete room/object categories disambiguate most referring expressions across scenes; yet the larger NE (4.54/5.48~m vs.~4.35/5.18~m) and lower SPL (0.47/0.39 vs.~0.50/0.43) show that texture-free semantics miss fine-grained details needed for precise goal alignment. The three-modal NavOne (RGB + occupancy + semantics) attains the best SR/SPL/NE on both splits and demonstrates why joint multi-modal fusion in the \emph{Top-Down Map Fuser} is required for robust TD-VLN.

%%%%%%%%%%%%%%%%%%%%%%%%%%%%%%%%%%%%%%%%%%%%%%%%%%%%%%%%%%%%%%%%%%%%%%%%%%%%%%%
%%%%%%%%%%%%%%%%%%%%%%%%%%%%%%%%%%%%%%%%%%%%%%%%%%%%%%%%%%%%%%%%%%%%%%%%%%%%%%%

\end{document}